\documentclass[10pt,twocolumn,letterpaper]{article}

\usepackage{subcaption}

\usepackage{3dv}
\usepackage{times}
\usepackage{epsfig}
\usepackage{graphicx}
\usepackage{amsmath}
\usepackage{amssymb}

\makeatletter
\@namedef{ver@everyshi.sty}{}
\makeatother

\usepackage{bm}
\usepackage{booktabs}
\usepackage{enumitem}
\usepackage{etoolbox}
\usepackage{newunicodechar}
\usepackage{mathtools}
\usepackage{multirow}
\usepackage[final]{pdfpages}
\usepackage{siunitx}
\usepackage{stfloats}  
\usepackage{tabu}
\usepackage{xcolor}

\usepackage{tikz}

\usepackage{pgfplots}   
\pgfplotsset{compat=newest}
\usepgfplotslibrary{groupplots}

\usepackage[pagebackref=false,breaklinks=true,letterpaper=true,colorlinks,bookmarks=false]{hyperref}

\threedvfinalcopy 

\def\threedvPaperID{213} 

\ifthreedvfinal\fi
\definecolor{githubColor}{HTML}{2EA44F}
\newcommand{\gitref}[2]{\href{#1}{\color{githubColor}{#2}}}%

\definecolor{newGray}{HTML}{808080}
\newcommand{\makegray}[1]{\color{newGray}#1}%

\newcolumntype{O}[1]{S[detect-weight, mode=text, table-format=#1]}

\definecolor{colorCircle}{HTML}{0072BD}
\DeclareRobustCommand\sensorcircle{\tikz \fill[black, fill=colorCircle] circle (0.75ex);}

\definecolor{colorRect}{HTML}{D95319}
\DeclareRobustCommand\sensorrect{\tikz \fill[black, fill=colorRect] rectangle (1.5ex, 1.5ex);}

\graphicspath{{images/}}

\hypersetup{colorlinks=false}

\renewcommand{\bfseries}{\fontseries{b}\selectfont} 
\robustify\bfseries             
\newrobustcmd{\B}{\bfseries} 

\newcommand{\method}[1]{\textsc{#1}}

\makeatletter
\def\thanks#1{\protected@xdef\@thanks{\@thanks
        \protect\footnotetext{#1}}}
\makeatother

\newcommand\copyrighttext{\footnotesize \textcopyright~2021 IEEE. Personal use of this material is permitted. Permission from IEEE must be obtained for all other uses, in any current or future media, including reprinting/republishing this material for advertising or promotional purposes, creating new collective works, for resale or redistribution to servers or lists, or reuse of any copyrighted component of this work in other works.
DOI: \href{https://doi.org/10.1109/3DV53792.2021.00035}{10.1109/3DV53792.2021.00035}
}

\newcommand\copyrightnotice{%
    \begin{tikzpicture}[remember picture,overlay]%
 	\node[anchor=south, xshift=-8pt, yshift=20pt] at (current page.south)%
 	{\fbox{\parbox{\dimexpr\textwidth-\fboxsep-\fboxrule\relax}{\copyrighttext}}};%
 	\end{tikzpicture}%
}

\newcommand{\mbeq}{\overset{!}{=}}

\newcommand{\mat}[1]{\boldsymbol{#1}}
\newcommand{\quat}[1]{\mathrm{#1}}
\renewcommand{\vec}[1]{\boldsymbol{#1}}

\newcommand{\mneg}{^\text{\rmfamily \textup{-}}}
\newcommand{\mpos}{^\text{\rmfamily \textup{+}}}
\newcommand{\norm}[1]{\left\lVert#1\right\rVert}
\newcommand{\vmnorm}[1]{\ensuremath \left\| #1 \right\|}
\newcommand{\trans}{^\text{\rmfamily \textup{T}}}

\newcommand{\nullspace}{\operatorname{null}}
\newcommand{\vecspan}{\operatorname{span}}
\newcommand{\vectorize}{\operatorname{vec}}

\newcommand{\sol}[1]{\hat{#1}}

\newcommand{\pspace}{\,}  

\begin{document}

\title{Globally Optimal Multi-Scale Monocular Hand-Eye Calibration\\Using Dual Quaternions}

\author{Thomas Wodtko$^{*}$ 
\quad Markus Horn$^{*}$ 
\quad Michael Buchholz 
\quad Klaus Dietmayer \\
Ulm University, Institute of Measurement, Control and Microtechnology\\ Albert-Einstein-Allee 41, 89081 Ulm, Germany \\
{\tt\footnotesize \{firstname\}.\{lastname\}@uni-ulm.de}
\thanks{
This work was supported by the State Ministry of Economic Affairs Baden-Württemberg (project U-Shift\,II, AZ\,3-433.62-DLR/60) and the Federal Ministry of Education and Research (BMBF) (project UNICARagil, FKZ\,16EMO0290).
}
\thanks{$^{*}$ \textit{Thomas Wodtko and Markus Horn are co-first authors. Corresponding author: Thomas Wodtko.}}
}

\maketitle
\thispagestyle{empty}

\begin{abstract}
In this work, we present an approach for monocular hand-eye calibration from per-sensor ego-motion based on dual quaternions.
Due to non-metrically scaled translations of monocular odometry, a scaling factor has to be estimated in addition to the rotation and translation calibration.
For this, we derive a quadratically constrained quadratic program that allows a combined estimation of all extrinsic calibration parameters.
Using dual quaternions leads to low run-times due to their compact representation.
Our problem formulation further allows to estimate multiple scalings simultaneously for different sequences of the same sensor setup.
Based on our problem formulation, we derive both, a fast local and a globally optimal solving approach.
Finally, our algorithms are evaluated and compared to state-of-the-art approaches on simulated and real-world data, e.g., the EuRoC MAV dataset.
\end{abstract}

\section{Introduction}

\copyrightnotice

Cameras are widely used in automated environments, such as autonomous vehicles or robotic systems, since they are both, cost and space efficient~\cite{sun2020scalability}.
Due to extensive research in this field, reliable and robust detectors and classifiers are available~\cite{schoen2021mgnet}.
However, a fundamental property of cameras is the projection of a 3D scene into a 2D representation.
This projection allows for efficient processing, but depth information is lost.
Therefore, without additional information, all motion estimates on monocular cameras are only up to an unknown scaling factor for the translation \cite{schmidt2005calibration}.
Only when using multiple sensors, e.g., in a stereo camera setup or in a setup with lidar and camera, available data can be combined to recover the lost depth information and the scaling factor.
In contrast, the task of monocular hand-eye calibration is to estimate the translation scaling in addition to the extrinsic calibration only from per-sensor ego-motion.
This makes monocular hand-eye calibration more flexible and suitable for any sensor setup, as long as the ego-motion of all sensors can be estimated.

For any multi-sensor setup, extrinsic parameters are fundamental, i.e., the transformation between two sensors is crucial for an accurate fusion.
Due to the projection property of monocular cameras, special targets are mostly used for their extrinsic calibration.
These engineered objects with known geometric properties are often designed to be easily detectable.
The checkerboard pattern is a common choice for such calibration targets \cite{furgale2013unified, tsai1989new}.
Given the known target size, depth information can be recovered during the calibration procedure.
However, target-based calibration normally requires a separate calibration procedure since the targets are not necessarily available within the usual work environment of, e.g., autonomous vehicles.
Thus, validating given calibrations or updating a calibration in an online manner using target-based calibration approaches is either cumbersome or not possible.
In contrast, hand-eye calibration is possible in unstructured environment, enabling an effortless calibration using ordinary motion sequences.

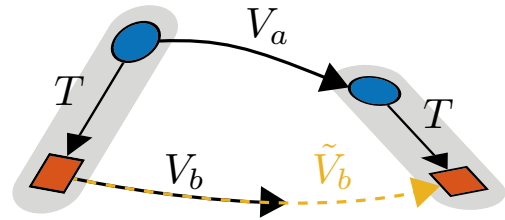
\begin{figure}
    \centering
    \resizebox{0.8\columnwidth}{!}{%
\begingroup%
  \makeatletter%
  \providecommand\color[2][]{%
    \errmessage{(Inkscape) Color is used for the text in Inkscape, but the package 'color.sty' is not loaded}%
    \renewcommand\color[2][]{}%
  }%
  \providecommand\transparent[1]{%
    \errmessage{(Inkscape) Transparency is used (non-zero) for the text in Inkscape, but the package 'transparent.sty' is not loaded}%
    \renewcommand\transparent[1]{}%
  }%
  \providecommand\rotatebox[2]{#2}%
  \newcommand*\fsize{\dimexpr\f@size pt\relax}%
  \newcommand*\lineheight[1]{\fontsize{\fsize}{#1\fsize}\selectfont}%
  \ifx\svgwidth\undefined%
    \setlength{\unitlength}{118.25025558bp}%
    \ifx\svgscale\undefined%
      \relax%
    \else%
      \setlength{\unitlength}{\unitlength * \real{\svgscale}}%
    \fi%
  \else%
    \setlength{\unitlength}{\svgwidth}%
  \fi%
  \global\let\svgwidth\undefined%
  \global\let\svgscale\undefined%
  \makeatother%
  \begin{picture}(1,0.45211106)%
    \lineheight{1}%
    \setlength\tabcolsep{0pt}%
    \put(0,0){\includegraphics[width=\unitlength,page=1]{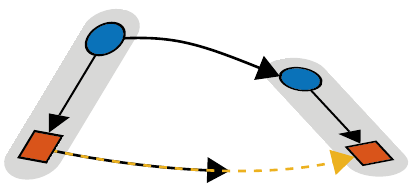}}%
    \put(0.31185599,0.07598481){\makebox(0,0)[lt]{\lineheight{1.25}\smash{\begin{tabular}[t]{l}$V_b$\end{tabular}}}}%
    \put(0.47453999,0.3666734){\makebox(0,0)[lt]{\lineheight{1.25}\smash{\begin{tabular}[t]{l}$V_a$\end{tabular}}}}%
    \put(0.82891406,0.1856647){\makebox(0,0)[lt]{\lineheight{1.25}\smash{\begin{tabular}[t]{l}$T$\end{tabular}}}}%
    \put(0.09342157,0.23218045){\makebox(0,0)[lt]{\lineheight{1.25}\smash{\begin{tabular}[t]{l}$T$\end{tabular}}}}%
    \put(0.60806697,0.07598481){\color[rgb]{0.9254902,0.69411765,0.12156863}\makebox(0,0)[lt]{\lineheight{1.25}\smash{\begin{tabular}[t]{l}$\tilde{V}_b$\end{tabular}}}}%
  \end{picture}%
\endgroup%
    }
    \caption{
    The transformation graph for two sensors, sensor $a$ (\sensorcircle) and sensor $b$ (\sensorrect), at two consecutive time steps is shown.
    For this motion, the unknown calibration $T$, the measured sensor motions $V_a$ and $V_b$, and the true metric-scaled motion $\tilde{V}_b$ are illustrated.
    }
    \label{fig:trafoCycle}
\end{figure}

Alternatively, by design, motion-based extrinsic hand-eye calibration does not require targets.
Instead, only per-sensor ego-motion estimates must be available, which are then used to retrieve the calibration.
This allows the calibration of a variety of different sensor types,
e.g., Inertial Measurement Units (IMUs), which in fact cannot detect calibration targets.
For sensors with metric-scaled motion estimation, hand-eye calibration has been extensively explored with many different approaches to choose from \cite{daniilidis1999hand, horaud1995hand, horn2021online, wise2020certifiably}.
However, the projection property of monocular cameras leads to an unknown scaling of their estimated translation motion.
This implies that the scaling has to be estimated in addition to the translation and rotation calibration when monocular cameras are involved.
Figure~\ref{fig:trafoCycle} shows the underlying transformation graph for monocular hand-eye calibration with the measured motion $V_b$ and the scaled motion $\tilde{V}_b$.
Furthermore, in case two monocular cameras are calibrated, only the relative scaling between them can be estimated without further data.
Although it is possible to use calibration targets for obtaining metric-scaled motion, this would lead to a combination of target and motion-based calibration.
However, this is contradicting with our objective of a fast calibration within unstructured environments.

For motion-based calibration, the chosen transformation representation is important for formulating the optimization problem.
Generally, homogeneous matrices (HMs) or dual quaternions (DQs) are widely used.
In contrast to HMs, rotation and translation are not considered separately with DQs.
Instead, the translation is represented in combination with the rotation in the dual part.
This complicates the integration of the translation scaling.
Nevertheless, it seems promising to formulate the scaled calibration also with DQs since we have already proven that the use of DQs for motion-based calibration without scaling is more efficient than the use of HMs \cite{horn2021online}.

Hence, we propose
\begin{itemize}[noitemsep,topsep=0pt]
    \item a unified Quadratically Constrained Quadratic Program (QCQP) formulation for non-metrically scaled motion-based calibration using dual quaternions in Section~\ref{sec:cost};
    \item an extension for multi-scale calibration in Section~\ref{sec:multi_scaling};
    \item a globally optimal and a fast local solving approach in Section~\ref{sec:algorithms};
    \item an extensive evaluation on real-world data from public datasets with artificial variations in Section~\ref{sec:experiments}; and
    \item a Python-based open-source library\footnote{\gitref{https://github.com/uulm-mrm/motion_based_calibration}{https://github.com/uulm-mrm/motion\_based\_calibration}} with our proposed calibration methods.
\end{itemize}

\section{Related Work}

Classic hand-eye calibration from metric-scaled motion data has already been widely covered in literature, mostly based on HMs \cite{giamou2019certifiably, horaud1995hand, ruland2012globally} or DQs \cite{brookshire2013extrinsic, daniilidis1999hand, horn2021online}.
Especially the globally optimal approaches \cite{giamou2019certifiably, horn2021online, ruland2012globally} yield accurate results, even on noisy data.
This work builds on our previous publication~\cite{horn2021online}, in which the DQ formulation leads to a less complex optimization problem and, thus, a faster optimization.

As stated before, when using hand-eye calibration with monocular cameras, the relative scale between the translations of both sensors has to be estimated in addition to the calibration.
When using engineered calibration targets \cite{furgale2013unified, tsai1989new}, the motion can be estimated with a metric scale.
This makes it possible to choose any of the classic hand-eye calibration approaches described before.
However, since our goal is to perform hand-eye calibration in unstructured environments, we cannot rely on specific, known targets within our environment.

Especially for visual-inertial odometry \cite{huang2020online, qin2018vins, yang2016monocular}, the correct scaling of monocular odometry is crucial for fusing visual and inertial measurements.
With known visual-inertial calibration, it is possible to estimate the scaling directly \cite{qin2018vins}, whereas for uncalibrated setups, scaling and calibration must be estimated simultaneously \cite{huang2020online, yang2016monocular}.

The previously mentioned methods for monocular hand-eye calibration have been developed for specific sensor types, in most cases cameras and IMUs.
In contrast, our focus is to provide a generic method for hand-eye calibration of arbitrary sensor types with possible scaling ambiguities that only relies on per-sensor ego-motion.
Most existing approaches for this use either HMs or DQs for representing transformations.
The usual derivations for HMs or DQs lead to two equations: the first is constraining the rotation only, and the second is jointly constraining rotation, translation, and scale.
A common approximation for reducing the complexity of the resulting optimization problem is to estimate the rotation from the first equation only and afterwards the translation and scale from the second equation \cite{heller2011structure, heng2013camodocal, wei2018calibration}.
However, as already observed by Horaud et al.~\cite{horaud1995hand}, for noisy measurements, this approximation leads to a greater error than a simultaneous estimation of all parameters.
Thus, for the best possible result on noisy data, this approximation is avoided in the following approaches as well as in our work.

For HMs, Wise et al.~\cite{wise2020certifiably} describe a certifiable globally optimal algorithm based on the linear formulation of Andreff et al.~\cite{andreff2001robot}.
They derive a QCQP and obtain a globally optimal solution through the Lagrangian dual problem and Semidefinite Programming (SDP).
For HMs and DQs, Schmidt et al.~\cite{schmidt2005calibration} derive two highly nonlinear cost functions.
However, as our experiments in Section~\ref{sec:experiments} show, their calibration performance varies greatly, depending on the dataset used.  
To the best of our knowledge, there is no method that combines the efficient DQ formulation and certifiable global optimality for hand-eye calibration with scaling.
Our work aims to fill this gap.

Furthermore, of all previously mentioned approaches, only \cite{heng2013camodocal} supports multiple scaling factors for different sequences of the same sensor setup.
With our problem formulation, we can include multiple scalings while still maintaining the QCQP form and solving rotation and translation simultaneously.

\section{Problem Formulation}
\label{sec:propForm}

In case of non-metrically scaled motion estimates, an additional scaling factor must be estimated during calibration.
For HMs, Wise et al.~\cite{wise2020certifiably} have proposed an extension of Giamou et al.~\cite{giamou2019certifiably} that handles the scaling as an additional optimization parameter.
Similarly, our DQ based approach is derived in the following, extending our previous approach for calibration with metric-scaled motion~\cite{horn2021online}.

\subsection{Notation}

We denote generic transformations without a fixed representation as functions $V$.
(Dual) quaternions are denoted with upright letters $\quat{q}$ and the respective vector representation with $\vec{q} = \vectorize{(\quat{q})}$.
The transformation chain $V_a \circ V_b$ represented as dual quaternions yields the multiplication $\quat{q}_a \quat{q}_b$.

Further, $\quat{q}_c = \quat{q}_a \quat{q}_b$ can be represented vectorized as matrix-vector products $\vec{q}_c = \mat{Q}\mpos_a \vec{q}_b = \mat{Q}\mneg_b \vec{q}_a$ with 
\begin{equation}
    \mat{Q}\mpos_a = \begin{bmatrix}
        \mat{R}\mpos_a & \mat{0}\\
        \mat{D}\mpos_a & \mat{R}\mpos_a
    \end{bmatrix} \pspace , \quad
    \mat{Q}\mneg_b = \begin{bmatrix}
        \mat{R}\mneg_b & \mat{0}\\
        \mat{D}\mneg_b & \mat{R}\mneg_b
    \end{bmatrix} \pspace ,
\end{equation}
where $\mat{R}\mpos_a$, $\mat{R}\mneg_b$ denote the respective real and $\mat{D}\mpos_a$, $\mat{D}\mneg_b$ the respective dual quaternion matrix representations \cite{mccarthy1990introduction}.

\subsection{Deriving a Cost Function}
\label{sec:cost}
 
First, we describe the implication of scaling a translation vector $\mat{t}$ with the factor $\alpha \in \mathbb{R}_{\ast}^{+}$ in the dual quaternion representation.
An ordinary dual quaternion is denoted as
\begin{equation}
\label{eq:dualQuaternion}
    \quat{q} = \quat{r} + \epsilon \, \quat{d} \pspace ,
\end{equation}
with the real part $\quat{r}$, the dual part $\quat{d}$, and the dual unit $\epsilon$~\cite{mccarthy1990introduction}.
For representing a transformation with rotation $\quat{r}$ and translation $\vec{t}$, the real part is $\quat{r}$ and the dual part is
\begin{equation}
\label{eq:dualQuatDualPart}
    \quat{d} = \tfrac{1}{2} \quat{t} \quat{r} \pspace , \quad \text{with\ } \quat{t} = (0,\vec{t}) \pspace .
\end{equation}
The scaled translation $\tilde{\mat{t}} = \alpha \, \mat{t}$ results in a scaled translation quaternion $\tilde{\quat{t}} = \alpha \, \quat{t}$.
Inserting $\tilde{\quat{t}}$ into~\eqref{eq:dualQuatDualPart} and~\eqref{eq:dualQuaternion} yields the scaled dual quaternion
\begin{equation}
    \tilde{\quat{q}} = \quat{r} + \epsilon \, \alpha \, \quat{d} \pspace.
\end{equation}
Due to the coupling property of DQs, not only the translation but also the rotation of the dual part is thereby scaled.

For the derivation of the optimization problem, no noise is considered.
Therefore, we assume w.l.o.g. that the translation of the second transformation $V_b$ is scaled.
However, under the influence of noise, the optimization result varies if $V_a$ is considered to be scaled instead. 
We will discuss this in Section~\ref{sec:noiseInfluence}.

The transformation cycle $V_a \circ T = T \circ \tilde{V}_b$, illustrated in Figure~\ref{fig:trafoCycle}, expressed with dual quaternions yields 
\begin{subequations}
\begin{gather*}
    \quat{q}_a \quat{q}_T =  \quat{q}_T \tilde{\quat{q}}_b \\
    \begin{aligned}
\iff  0 = &\pspace \quat{r}_a \quat{r}_T - \quat{r}_T \quat{r}_b \\
          &   + \varepsilon \, (\quat{d}_a \quat{r}_T + \quat{r}_a \quat{d}_T - \quat{d}_T \quat{r}_b - \alpha \, \quat{r}_T \quat{d}_b) \pspace . 
    \end{aligned}
\end{gather*}
\end{subequations}
Substituting the dual quaternion multiplication with its respective matrix-vector representation and using $\vec{r} = \vectorize{(\quat{r}_T)}$, the real part can be expressed as 
\begin{subequations}
\begin{align}
\vec{0} & \pspace = \pspace \vectorize{(\quat{r}_a \quat{r}_T - \quat{r}_T \quat{r}_b)} \\
  & \pspace = \pspace \mat{R}\mpos_a \mat{r} - \mat{R}\mneg_b \mat{r} \\
  & \pspace = \pspace (\mat{R}\mpos_a- \mat{R}\mneg_b) \mat{r}\pspace.
\label{eq:realPart}
\end{align}
\end{subequations}
Analogously, with $\vec{d} = \vectorize{(\quat{d}_T)}$ and the substitution $\mat{s} = \alpha \, \mat{r}$, the dual part yields
\begin{subequations}
\begin{align}
\vec{0} & = \pspace \vectorize{(\quat{d}_a \quat{r}_T + \quat{r}_a \quat{d}_T 
        - \quat{d}_T \quat{r}_b - \alpha \, \quat{r}_T \quat{d}_b)} \\
  & = \pspace \mat{D}\mpos_a \mat{r}  + \mat{R}\mpos_a \mat{d}
        - \mat{R}\mneg_b \mat{d} - \alpha \, \mat{D}\mneg_b \mat{r} \\
  & = \pspace \mat{D}\mpos_a \mat{r} - \mat{D}\mneg_b \mat{s}  
        + (\mat{R}\mpos_a - \mat{R}\mneg_b) \mat{d} \pspace . 
\label{eq:dualPart}
\end{align}
\end{subequations}
Combining~\eqref{eq:realPart} and~\eqref{eq:dualPart} leads to
\begin{equation}
\label{eq:Mx}
\mat{M} \Vec{x} :=
\begin{bmatrix}
\mat{R}\mpos_a - \mat{R}\mneg_b & \mat{0} & \mat{0} \\
\mat{D}\mpos_a & - \mat{D}\mneg_b & \mat{R}\mpos_a - \mat{R}\mneg_b
\end{bmatrix}
\begin{bmatrix}
\mat{r} \\ \mat{s} \\ \mat{d}
\end{bmatrix}
=
\mat{0} \pspace .
\end{equation}

For $n$ motion pairs, the matrix $\mat{M}$ in \eqref{eq:Mx} of each step $t$ is denoted by $\mat{M}_t$.
Applying the quadratic norm to~\eqref{eq:Mx} and adding up all steps results in the cost function
\begin{subequations}
\begin{gather}
    J(\vec{x}) = \!\sum_{t=1}^{n} \vec{x}\trans \mat{M}_t\trans \mat{M}_t \vec{x} = \vec{x}\trans \mat{Q} \vec{x} \pspace ,\\
\text{with\ } \mat{Q} := \!\sum_{t=1}^{n} \mat{M}_t\trans \mat{M}_t \pspace .
\end{gather}
\end{subequations}

Similar to~\cite{horn2021online}, the constraint $\vec{g}_{d}(\vec{x}) \mbeq 0$ is required to ensure a valid unit dual quaternion:
\begin{equation}
\label{eq:dualConst}
\vec{g}_{d}(\vec{x}) 
= 
\begin{bmatrix}
1 - \norm{\vec{r}}^2_2 \\
2 \vec{r}\trans \vec{d} \\
\end{bmatrix}  
\mbeq
\vec{0} \pspace .
\end{equation}
Further, additional constraints are necessary to enforce the substitution $\vec{s} = \alpha \, \vec{r}$ to hold during optimization.
Given that $\exists i,\, r_i \neq 0$ due to $\norm{\vec{r}}_2 = 1$, the six necessary constraints are defined by
\begin{equation}
\label{eq:alpha}
    r_i s_j - r_j s_i = 0 \pspace , \quad \text{with } (i, j) \in \{1, \dots, 4\}^2, \, i < j \pspace .
\end{equation}
The constraints are explained in more detail in the supplementary material.

In order to reduce the computational complexity, we neglect the special case $r_1 = 0$, which only occurs if the rotation angle between the sensors is exactly $180^{\circ}$.
This leads to only three necessary equality constraints for our problem:
\begin{equation}
\label{eq:alphaConst}
\vec{g}_{\alpha}(\vec{x}) 
= 
\begin{bmatrix}
r_1 s_2 &-& r_2 s_1 \\
r_1 s_3 &-& r_3 s_1 \\
r_1 s_4 &-& r_4 s_1
\end{bmatrix}  
\mbeq
\vec{0} \pspace .
\end{equation}
Given the solution $\sol{\vec{x}}$ of the optimization problem, for which $\norm{\sol{\vec{r}}}_2 = 1$ holds, the estimated scaling $\sol{\alpha}$ is calculated by $\sol{\alpha} = \norm{\sol{\vec{s}}}_2$.

Extending the optimization problem of~\cite{horn2021online} with the scaled rotation $\vec{s}$ and the additional constraints $\vec{g}_{\alpha}(\vec{x})$, the resulting optimization problem for a scaled extrinsic calibration is given by
\begin{subequations}
\label{eq:optiProb}
\begin{align}
    \!\min_{\vec{x}\in\mathbb{R}^{12}} & \quad J(\vec{x}) \\
    \text{w.r.t.} & \quad \vec{g}_{d}(\vec{x}) \mbeq \vec{0} \, \land \, \vec{g}_{\alpha}(\vec{x}) \mbeq \vec{0} \pspace .
\end{align}
\end{subequations}
Due to the constraints, the optimization problem~\eqref{eq:optiProb} is non-convex. 
Therefore, the Lagrangian dual problem is derived in the following for a globally optimal solution.

\subsection{Lagrangian Dual Problem}
\label{sec:lagDualProb}

The Lagrangian dual problem for optimization problem \eqref{eq:optiProb} is derived in this section.
Similar to~\cite{giamou2019certifiably, horn2021online, wise2020certifiably}, the resulting Lagrangian dual problem is an SDP problem since the primal problem has a QCQP form.
In contrast to~\cite{horn2021online}, the additional constraints~\eqref{eq:alphaConst} are taken into account in~\eqref{eq:optiProb} for monocular calibration.

Given the Lagrange function of~\eqref{eq:optiProb}
\begin{equation}
    L(\vec{x},\vec{\lambda}) = \vec{x}\trans\mat{Q}\vec{x} + \vec{\lambda_d}\trans \, \vec{g}_{d}(\vec{x}) + \vec{\lambda_{\alpha}}\trans \, \vec{g_{\alpha}}(\vec{x}) \pspace ,
\end{equation}
with $\vec{\lambda_d} \in \mathbb{R}^{2}$ and $\vec{\lambda_{\alpha}} \in \mathbb{R}^{3}$, all constraints are first represented in a quadratic manner.
With
\begin{equation}
    \label{eq:lagrange_constraints}
    \mat{P}_{d,1} = 
    \begin{bmatrix}
        - \textbf{I}_{4 \times 4} & \vec{0}_{4 \times 8}\\
        \vec{0}_{8 \times 4} & \vec{0}_{8 \times 8}
    \end{bmatrix} , \;
    \mat{P}_{d,2} = 
    \begin{bmatrix}
        \vec{0}_{4} & \textbf{I}_{4} &\vec{0}_{4} \\
        \textbf{I}_{4} & \vec{0}_{4} &\vec{0}_{4} \\
        \vec{0}_{4} & \vec{0}_{4} &\vec{0}_{4} \\
    \end{bmatrix} ,
\end{equation}
and $\mat{P}_{\alpha,1}, \mat{P}_{\alpha,2},$ and $\mat{P}_{\alpha,3}$ satisfying
\begin{equation}
\label{eq:PFromAlpha}
\vec{x}\trans \, \mat{P}_{\alpha,i} \, \vec{x} = g_{\alpha,i}(\vec{x}) \pspace , i\in\{1,2,3\} \pspace ,
\end{equation}
the Lagrange function can be expressed by
\begin{equation}
    \label{eq:LagFunc}
    L(\vec{x},\vec{\lambda}) = \vec{x}\trans\mat{Z}(\vec{\lambda)}\vec{x} + \lambda_1
    \pspace ,
\end{equation}
where $\mat{Z}$ combines all quadratic parts:
\begin{align}
    \mat{Z}(\vec{\lambda})  = \mat{Q} & + \lambda_1 \mat{P}_{d,1} + \lambda_2 \mat{P}_{d,2} \nonumber\\
    & + \lambda_3 \mat{P}_{\alpha,1} + \lambda_4 \mat{P}_{\alpha,2} + \lambda_5 \mat{P}_{\alpha,3} \pspace .
\end{align}
Similar to~\cite{horn2021online}, the Lagrangian dual problem is subsequently given by
\begin{subequations}
\label{eq:lagDualProb}
\begin{align}
\!\max_{\lambda_1} & \quad \Theta(\vec{\lambda}) := \lambda_1\\
\text{w.r.t.}      & \quad \mat{Z}(\vec{\lambda}) \succeq \mat{0} \pspace .
\end{align}
\end{subequations}
As mentioned before, \eqref{eq:lagDualProb} belongs to the group of SDP problems, for which solvers like~\cite{agrawal2018rewriting, diamond2016cvxpy} are publicly available.

\subsection{Multi-Scale Calibration}
\label{sec:multi_scaling}

When using multiple sequences for calibrating the same sensor pair, each sequence might have a different scaling.
This also happens when the ego-motion estimation of a scaled sensor is reinitialized, for example, due to kidnapping \cite{heng2013camodocal}.

For multiple scalings $\alpha_j$ with $j = 1, \dots, m$, the variables in \eqref{eq:Mx} are replaced by
\begin{subequations}
\begin{gather}
\mat{x} :=
\begin{bmatrix}
\mat{r} & \mat{s}_{1} & \dots & \mat{s}_{m} & \mat{d}
\end{bmatrix}\trans \\[\medskipamount]
\label{eq:MxMultiScale}
\mat{M}_{j} :=
\begin{bmatrix}
\mat{R}\mpos_a - \mat{R}\mneg_b & \mat{0} & \dots & \mat{0} & \mat{0} \\
\mat{D}\mpos_a & \mat{S}_1 & \dots & \mat{S}_{m} & \mat{R}\mpos_a - \mat{R}\mneg_b
\end{bmatrix} \pspace , \\
\text{with\ }\mat{S}_k =
\begin{cases}
    -\mat{D}\mneg_b \pspace , & \text{if } k = j\\
    \mat{0} \pspace , & \text{else} 
\end{cases} \pspace ,
\end{gather}
\end{subequations}
using the substitutions $\mat{s}_j = \alpha_j \mat{r}$.
With $n_j$ steps for each scaling, the new cost function is then given by
\begin{subequations}
\begin{gather}
    J(\vec{x}) = \vec{x}\trans \mat{Q} \vec{x} \pspace , \\
\text{with\ } \mat{Q} := \sum_{j=1}^{m} \sum_{t=1}^{n_j} \mat{M}_{j, t}\trans \mat{M}_{j, t} \pspace ,
\end{gather}
\end{subequations}
with the matrix $M_{j, t}$ for each step $t$ of the scaling $j$.
Analogously to the single-scale case, three additional optimization constraints are necessary for each substitution:
\begin{equation}
\label{eq:alphaConstMulti}
\vec{g}_{\alpha_j}(\vec{x}) 
= 
\begin{bmatrix}
r_1 s_{j,2} &-& r_2 s_{j,1} \\
r_1 s_{j,3} &-& r_3 s_{j,1} \\
r_1 s_{j,4} &-& r_4 s_{j,1}
\end{bmatrix}  
=
\vec{0} \pspace .
\end{equation}
Subsequently, the optimization problem can be set up similarly to the single-scale case, formally given by
\begin{subequations}
\label{eq:optiProbMulti}
\begin{align}
    \!\min_{\vec{x}\in\mathbb{R}^{8+4m}} & \quad J(\vec{x}) \\
    \text{w.r.t.}	 & \quad \vec{g}_{d}(\vec{x}) \mbeq \vec{0} \pspace ,\\
    & \quad \vec{g}_{\alpha_j}(\vec{x}) \mbeq \vec{0} \pspace , \quad \text{for } j = 1, \dots, m
    \pspace .
\end{align}
\end{subequations}

The Lagrangian dual problem can be derived for the multi-scale case analogously to Section~\ref{sec:lagDualProb}.
To do so, further constraints, represented by the matrices $\mat{P}_{\alpha_j,1}, \mat{P}_{\alpha_j,2},$ and $\mat{P}_{\alpha_j,3}$, must be added for each scaling factor $\alpha_j$.
Given the additional constraints $\vec{g}_{\alpha_j}(\vec{x})$, the matrices can be set up similarly to~\eqref{eq:PFromAlpha}.
Consecutively, the Lagrange function is given by replacing $\mat{Z}$ in~\eqref{eq:LagFunc} with
\begin{align}
    \mat{Z}(\vec{\lambda}) = \mat{Q} & + \lambda_1 \mat{P}_{d,1} + \lambda_2 \mat{P}_{d,2} \nonumber\\
    & + \sum_{j=1}^{m} \sum_{k=1}^{3}
        \lambda_{3j + k -1} \mat{P}_{\alpha_j,k} 
    \pspace.
\end{align}
Finally, the Lagrangian dual problem is given by~\eqref{eq:lagDualProb}, using the replaced $\mat{Z}$.
For a given $m$, the dimension of $\vec{\lambda}$ is $2 + 3 m$.

\subsection{Scaling Sensor a}

With $\mat{M}$ from \eqref{eq:Mx} or \eqref{eq:MxMultiScale}, the motion estimates of sensor~$b$ are always considered to be scaled.
However, if sensor~$a$'s motion should be scaled instead, $\mat{M}$ in \eqref{eq:Mx} can be replaced with
\begin{equation}
\label{eq:Max}
\mat{M} :=
\begin{bmatrix}
\mat{R}\mpos_a - \mat{R}\mneg_b & \mat{0} & \mat{0} \\
- \mat{D}\mneg_b & \mat{D}\mpos_a & \mat{R}\mpos_a - \mat{R}\mneg_b
\end{bmatrix} \pspace .
\end{equation}
This is likewise possible for the multi-scale case in \eqref{eq:MxMultiScale}.

In the noise-free case, the scaling of $a$ yields the inverse of the scaling of $b$.
However, we observed a significant performance difference on noisy real-world data, depending on which sensor motion is scaled.
An in-depth evaluation of this behavior is given in Section~\ref{sec:noiseInfluence}.

\section{Algorithms}

\label{sec:algorithms}
In this section, we derive a fast local and a globally optimal solving approach for the two optimization problems of Section~\ref{sec:propForm} based on the methods from~\cite{horn2021online}.
The approaches are later referred to as \method{Fast} and \method{Global} approach.

\subsection{Fast Calibration}

Given the optimization problem~\eqref{eq:optiProb} or \eqref{eq:optiProbMulti}, a Sequential Quadratic Programming (SQP) solver is used for a fast local optimization.
In~\cite{horn2021online}, this was proven to be better suited than Interior Point (IP) methods due to the quadratic form of the problem.
During the optimization, the constraints are enforced by the optimizer.

Although an SQP solver obtains local solutions only, the globality of a solution can be verified using our approach proposed in~\cite{horn2021online}.
Hereby, the dual variables $\vec{\lambda}$ to the local solution are estimated using the first-order optimality condition of the dual problem.
If the estimates $\sol{\vec{\lambda}}$ fulfill all dual optimization constraints, the globality is guaranteed.

\subsection{Globally Optimal Calibration}

In addition to the fast local approach, both optimization problems \eqref{eq:optiProb} or \eqref{eq:optiProbMulti} can also be solved in a global manner.
The respective Lagrangian dual problems have been derived in Section~\ref{sec:propForm}.
This type of problem can be efficiently solved using an SDP solver. 
However, these solvers only yield the globally optimal dual solution and the respective primal solution must be recovered in a consecutive step.

Given the globally optimal dual solution $\sol{\vec{\lambda}}$, the first order optimality constraint yields
\begin{equation}
    \frac{\partial L(\vec{x},\sol{\vec{\lambda}})}{\partial \vec{x}} = 2\mat{Z}(\sol{\vec{\lambda}})\vec{x} \stackrel{!}{=} \vec{0} \pspace .
\end{equation}
Therefore, since a globally optimal primal solution must exist for our problem, it must be within the null space of $\mat{Z}(\sol{\vec{\lambda}})$.
Assuming a one-dimensional null space, as in~\cite{giamou2019certifiably, horn2021online, wise2020certifiably}, a local primal solution can be uniquely recovered by using the first constraint of $\vec{g}_d$ with
\begin{align}
\label{eq:dualToPrimal}
\sol{\vec{x}} & = \frac{\vec{v}}{\vmnorm{\vec{v}_{\{1,...,4\}}}_2} \pspace ,
\end{align} 
where $\vec{v}$ must satisfy $\vecspan(\{\vec{v}\}) = \nullspace(\textbf{Z}(\sol{\vec{\lambda}}))$.
Due to the one-dimensional null space, enforcing a single constraint is sufficient to uniquely recover the primal solution $\sol{\vec{x}}$ that satisfies all primal constraints.
Given $\sol{\vec{x}}$, the duality gap is defined by
\begin{equation}
    J(\sol{\vec{x}}) - \Theta(\vec{\sol{\lambda}}) = \sol{\vec{x}}\trans \mat{Z}(\sol{\vec{\lambda}})\sol{\vec{x}} = 0\pspace ,
\end{equation}
with $J(\sol{\vec{x}})$ from \eqref{eq:optiProb} and $\Theta(\vec{\sol{\lambda}})$ from \eqref{eq:lagDualProb}.
Thus, the duality gap is ensured to be zero and the local primal solution $\sol{\vec{x}}$ is globally optimal.
In case the null space assumption does not hold, the global approach does not yield any solution.
Thus, a globally optimal solution cannot be guaranteed.
Nevertheless, if a solution is found, its globality is guaranteed post hoc.

\section{Experiments}
\label{sec:experiments}

In this section, we evaluate our proposed algorithms and discuss the results.
Additionally, the \method{Fast} and the \method{Global} algorithm are compared to other state-of-the-art approaches for monocular hand-eye calibration~\cite{schmidt2005calibration, wei2018calibration, wise2020certifiably}.
First, the influence of noise is evaluated on simulated data~\cite{wise2020certifiably}.
Second, all approaches are evaluated on a publicly available real-world dataset~\cite{brookshire2013extrinsic} with artificial scaling.
This makes it possible to compare the estimated scaling with a known ground-truth.
Afterwards, the publicly available EuRoC MAV dataset~\cite{burri2016euroc} with real-world recordings of monocular cameras is used for comparison.

We use the same error metrics as described in~\cite{horn2021online}, covering physical entities, i.e., the rotation magnitude and translation offset.
Given a predicted calibration $(\sol{\quat{q}}, \sol{\alpha})$ and its respective ground truth calibration $(\quat{q}_T, \alpha)$, the errors in rotation, translation, and scaling are given by
\begin{subequations}
\begin{alignat}{2}
    \varepsilon_r &= 2 \arccos(\vec{q}_{\varepsilon,1}) \pspace ,\\
    \varepsilon_t &= \norm{2 \, \quat{q}_{\varepsilon,d} \, \quat{q}^*_{\varepsilon,r}} \pspace,\\
    \varepsilon_\alpha &= \left| \, \sol{\alpha} - \alpha \, \right| \pspace, 
\end{alignat}
\end{subequations}
with $\quat{q}_\varepsilon = \quat{q}_T^{-1} \, \sol{\quat{q}}$ and $\vec{q}_\varepsilon = \vectorize(\quat{q}_\varepsilon)$.

All compared approaches are implemented in Python using either SciPy~\cite{virtanen2020scipy} or CVXPY \cite{agrawal2018rewriting, diamond2016cvxpy} for optimization.
We have implemented the cost function for HMs and DQs of Schmidt et al.~\cite{schmidt2005calibration} and the DQ approach of Wei et al.~\cite{wei2018calibration}.
The HM approach of Wise et al.~\cite{wise2020certifiably} is evaluated using their provided source code.
As a baseline with unscaled calibration, we use the global approach of our previous work~\cite{horn2021online}.
In the following, these approaches are referred to as \method{Schmidt HM}, \method{Schmidt DQ}, \method{Wei}, \method{Wise}, and \method{Horn}, respectively.

All experiments were run on a general purpose computer equipped with an ADM\,Ryzen\textsuperscript{TM}\,7\,3700X CPU and 64GB of DDR4 RAM.
Given an optimization result, for some approaches, the calibration must be recovered from the result in further steps.
Thus, for a fair comparison, execution times include optimization and recovery times.
If not stated differently, timings are averaged over 100 runs.

\subsection{Asymmetric Influence of Noise}
\label{sec:noiseInfluence}

As a first step, we analyze the influence of noise on the simulated data provided by Wise et al.~\cite{wise2020certifiably}.
They provide 1000 transformation pairs with motion in all dimensions.
The use of simulated data makes it possible to define exact and known noise parameters for each sensor and ensures a well-conditioned optimization problem.
As mentioned in Section~\ref{sec:propForm}, for noisy measurements, we observed that the calibration performance depends on which sensor motion is assumed to be scaled.
For investigating this behavior, we added artificial noise to the metric-scaled data.
Thereby, noise parameters were chosen relative to the translation length and rotation magnitude of each step from $0-\SI{20}{\percent}$.
We created $10$ noisy datasets per parameter set.
\begin{figure}
\begin{subfigure}[a]{\columnwidth}
    \centering
    \resizebox{0.5\columnwidth}{!}{%
\begin{tikzpicture}

\pgfplotsset{every tick label/.append style={font=\LARGE}}
\begin{groupplot}[
group style={group size=1 by 1,horizontal sep=1.5cm, vertical sep=2cm},
]

\nextgroupplot[
colorbar,
colorbar style={ylabel={}},
colormap/viridis,
point meta max=1,
point meta min=0,
tick align=outside,
tick pos=both,
x grid style={white!70!black},
xlabel={Noise Sensor $b$ [\%]},
xmin=0, xmax=0.2,
xtick style={color=black},
xtick={0, 0.05, 0.1, 0.15, 0.2},
xticklabels={0, 5, 10, 15, 20},
y grid style={white!70!black},
ylabel={Noise Sensor $a$ [\%]},
ymin=0, ymax=0.2,
ytick style={color=black},
ytick={0, 0.05, 0.1, 0.15, 0.2},
yticklabels={0, 5, 10, 15, 20},
label style={font=\LARGE},
title= $\varepsilon_t$ [$\si{\metre}\text{]}$,
title style={font=\LARGE}
]
\addplot graphics [includegraphics cmd=\pgfimage,xmin=0, xmax=0.2, ymin=-0.002, ymax=0.2] {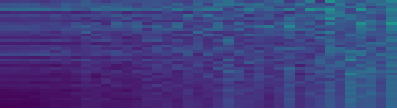};

\end{groupplot}

\end{tikzpicture}
    }
    \caption{Calibration without scaling \cite{horn2021online}.}
    \label{fig:noiseInfluenceUnscaled}
\end{subfigure}
\par\medskip %
\begin{subfigure}[b]{\columnwidth}
    \centering
    \resizebox{1\columnwidth}{!}{%
\begin{tikzpicture}

\pgfplotsset{every tick label/.append style={font=\LARGE}}
\begin{groupplot}[
group style={group size=2 by 1,horizontal sep=3.5cm, vertical sep=2cm},
]

\nextgroupplot[
colorbar,
colorbar style={},
colormap/viridis,
point meta max=1,
point meta min=0,
tick align=outside,
tick pos=both,
x grid style={white!70!black},
xlabel={Noise Sensor $b$ [\%]},
xmin=0, xmax=0.2,
xtick style={color=black},
xtick={0, 0.05, 0.1, 0.15, 0.2},
xticklabels={0, 5, 10, 15, 20},
y grid style={white!70!black},
ylabel={Noise Sensor $a$ [\%]},
ymin=0, ymax=0.2,
ytick style={color=black},
ytick={0, 0.05, 0.1, 0.15, 0.2},
yticklabels={0, 5, 10, 15, 20},
label style={font=\LARGE},
title= $\varepsilon_t$ [$\si{\metre}\text{]}$,
title style={font=\LARGE}
]
\addplot graphics [includegraphics cmd=\pgfimage,xmin=0, xmax=0.2, ymin=0, ymax=0.2] {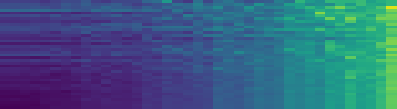};

            
\nextgroupplot[
colorbar,
colorbar style={
    ytick={0.0,0.417,0.833},
    yticklabels={0,0.05,0.1}
},
colormap/viridis,
point meta max=1,
point meta min=0,
tick align=outside,
tick pos=both,
x grid style={white!70!black},
xlabel={Noise Sensor $b$ [\%]},
xmin=0, xmax=0.2,
xtick style={color=black},
xtick={0, 0.05, 0.1, 0.15, 0.2},
xticklabels={0, 5, 10, 15, 20},
y grid style={white!70!black},
ymin=0, ymax=0.2,
ytick style={color=black},
ytick={0, 0.05, 0.1, 0.15, 0.2},
yticklabels={0, 5, 10, 15, 20},
label style={font=\LARGE},
title= $\varepsilon_\alpha$ [unitless\text{]},
title style={font=\LARGE}
]
\addplot graphics [includegraphics cmd=\pgfimage,xmin=0, xmax=0.2, ymin=0, ymax=0.2] {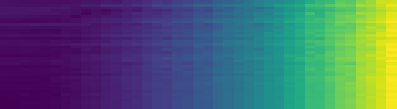};
\end{groupplot}

\end{tikzpicture}
    }
    \caption{\method{Global} calibration with scaling.}
    \label{fig:noiseInfluenceScaled}
\end{subfigure}
\caption{
    Results for the noise influence evaluation, comparing the baseline \method{Horn} (\subref{fig:noiseInfluenceUnscaled}) and our \method{Global} calibration (\subref{fig:noiseInfluenceScaled}), where sensor~$b$ is scaled.
    In contrast to the errors for an unscaled calibration, the errors for a scaled calibration are asymmetrically dependent on the noise levels of the sensors $a$ and $b$.
    Thus, the quality of the calibration result highly depends on which sensor is considered for scaling.
}
\label{fig:noiseInfluence}
\end{figure}

The median values of the results for the baseline method \method{Horn} without scaling as well as with our proposed \method{Global} method are displayed in Figure~\ref{fig:noiseInfluence}. 
For the regular calibration without scaling estimation, noise has a symmetric influence for both sensors on the calibration performance.
However, for the scaled calibration, the heatmap clearly shows that the influence of the noise is much higher for sensor $b$, which is assumed to be scaled.
The same behavior was also observed when swapping sensors $a$ and $b$ while still scaling the same sensor.
Furthermore, other methods for scaled calibration~\cite{schmidt2005calibration, wei2018calibration, wise2020certifiably} show the same behavior when swapping the sensors.

This leads to the conclusion that the scaled sensor motion should always be the motion with less noise.
For the following evaluation, we always consider this by swapping the sensors accordingly.
This is necessary, since in contrast to our method, the other methods were designed for scaling only one of the two sensor motions without providing the option for selecting the scaled motion.

\subsection{Artificial Scaling}
\label{sec:artScale}
\begin{table*}[t]
    \begin{center}
    \setlength{\tabcolsep}{0.37em}
    \begin{tabular}{lO{2.2}O{1.3}O{1.3}O{4.2}|O{2.2}O{1.3}O{1.3}O{4.2}|O{2.2}O{1.3}O{1.3}O{4.2}}
    \toprule
    \textbf{Method} & \multicolumn{4}{c}{\textbf{1x scale}} &
                        \multicolumn{4}{c}{\textbf{10x scale}} &
                        \multicolumn{4}{c}{\textbf{0.01x scale}} \\
    & $\varepsilon_t$ [$\si{\centi\metre}$] & $\varepsilon_r$ [$\si{\degree}$] & $\hat{\alpha}$ &{$t$ [$\si{\milli\second}$]}
    & $\varepsilon_t$ [$\si{\centi\metre}$] & $\varepsilon_r$ [$\si{\degree}$] & $\hat{\alpha}$ &{$t$ [$\si{\milli\second}$]}
    & $\varepsilon_t$ [$\si{\centi\metre}$] & $\varepsilon_r$ [$\si{\degree}$] & $\hat{\alpha}$ &{$t$ [$\si{\milli\second}$]}\\
    
    \midrule
    
    \method{Horn}~\cite{horn2021online} 
        & 1.10 & 1.049 & | & 5.39
        & 3.56 & 1.039 & | & 5.39
        & {\makegray{\raisebox{0.2em}{\tiny\textgreater}$10^2$}} & {\makegray{\raisebox{0.2em}{\tiny\textgreater}$10$}} & {\makegray{|}} & \makegray{5.67}\\

    \midrule
    
    \method{Schmidt HM}~\cite{schmidt2005calibration}
        & 1.10 & 1.070 & 0.985 & 4643.37
        & 1.31 & 1.070 & 9.85 & 5665.24
        & 1.31 & 1.070 & \B 0.010 & 5075.67\\
    
    \method{Schmidt DQ}~\cite{schmidt2005calibration}
        & 5.39 & 2.350 & 0.100 & 802.25
        & \makegray{5.32} & {\makegray{\raisebox{0.2em}{\tiny\textgreater}$90$}} & \makegray{0.00} & \makegray{1561.94}
        & \B 0.94 & 3.166 & \B 0.010 & 894.03\\
    
    \method{Wei}~\cite{wei2018calibration}
        & 1.11 & 1.070 & \B 0.997 & 9.12
        & 1.11 & 1.070 & \B 9.97 & \B 9.21 
        & 1.11 & 1.070 & \B 0.010 & 9.60\\
        
    \method{Wise}~\cite{wise2020certifiably}
        & 1.74 & 1.047 & 0.992 & 25.18
        & 1.74 & 1.047 & 9.92 & 24.79 
        & 1.74 & 1.047 & \B 0.010 & 25.29\\
        
    \method{Fast} (ours)
        & \B 1.08 & \B 0.931 & \B 0.998 & \B 5.32
        &    1.10 &    1.049 & \B 9.97 & 11.19
        &    1.08 & \B 0.933 & \B 0.010 & \B 5.39\\
    
    \method{Global} (ours)
        & \B 1.08 & \B 0.929 & \B 0.998 & 8.90
        & \B 1.07 & \B 0.927 & \B 9.98 & 11.13
        &    1.08 & \B 0.927 & \B 0.010 & 8.77 \\
    
    \bottomrule    
\end{tabular}

    \end{center}
    \caption{
        Results on the Brookshire dataset~\cite{brookshire2013extrinsic} are     shown. 
        The original motion for one sensor was artificially scaled by the given factor.
        \method{Horn} is not able to consider scaling and is only given as a baseline.
    }
    \label{tab:brookshire_scale}
\end{table*}

For our evaluation with artificial scaling, we use the real-world dataset provided by Brookshire et al.~\cite{brookshire2013extrinsic}.
They provide 200 interpolated pose estimates for two time-synchronized RGB-D cameras.
Since the cameras are not trigger-synchronized, the poses of sensor $b$ are interpolated at the timestamps of $a$.
Therefore, we assume sensor $a$ to be scaled, as mentioned in the previous section.
The use of RGB-D cameras leads to metric-scaled motion.
Thus, we applied artificial scaling with the scaling factors $10$ and $0.01$.
In combination with the metric-scaled data, this leads to three differently scaled, partially augmented datasets.
In order to demonstrate the importance of the scale estimation for monocular calibration, we have also evaluated the approach for unscaled hand-eye calibration~\cite{horn2021online} on this data as a baseline.

The results are presented in Table~\ref{tab:brookshire_scale}.
In general, all approaches except \method{Schmidt DQ} find valid solutions.
Furthermore, the performance of \method{Horn} decreases as expected for scaled data, which demonstrates the importance of the scaling estimation for the calibration with non-metrically scaled data.

With the exception of \method{Schmidt DQ}, the monocular calibration approaches can compensate for artificial scaling and recover $\alpha$ with only small discrepancies.
Both approaches of Schmidt et al.~\cite{schmidt2005calibration} are time-consuming with execution times up to multiple seconds, caused by their highly non-linear cost functions. 
In some cases, this even causes the optimization to be unfeasible.
Especially, \method{Schmidt DQ} does not yield good solutions for the scalings $1$ and $10$.
Compared to \method{Wise}, our approaches are noticeably faster and more accurate.
Although our problem dimension is higher, we have to enforce fewer constraints, which seems to have a greater impact on execution times.

\method{Wei} obtains similar results as our methods for the translation but is slightly less accurate for the rotation.
This is most likely caused by their separate solving of rotation and translation.
\method{Wei} and both of our approaches are the fastest.
Since our primal problem is not convex, the local solution of \method{Fast} can differ from the globally optimal solution of \method{Global}.
Nevertheless, the local solution is almost identical for the scalings $1$ and $0.01$.
This shows that \method{Fast} is a feasible choice for applications like online calibration, especially since the globality of the local solution can be verified.
If the execution time is of secondary importance, \method{Global} is the preferred method.
Summarizing this section, our proposed methods yield the best results for all scalings and are, at the same time, among the fastest approaches.

\subsection{Artificial Noise}
\label{sec:artNoise}

\begin{figure}[t]
    \centering
    \resizebox{0.91\columnwidth}{!}{%
        \input{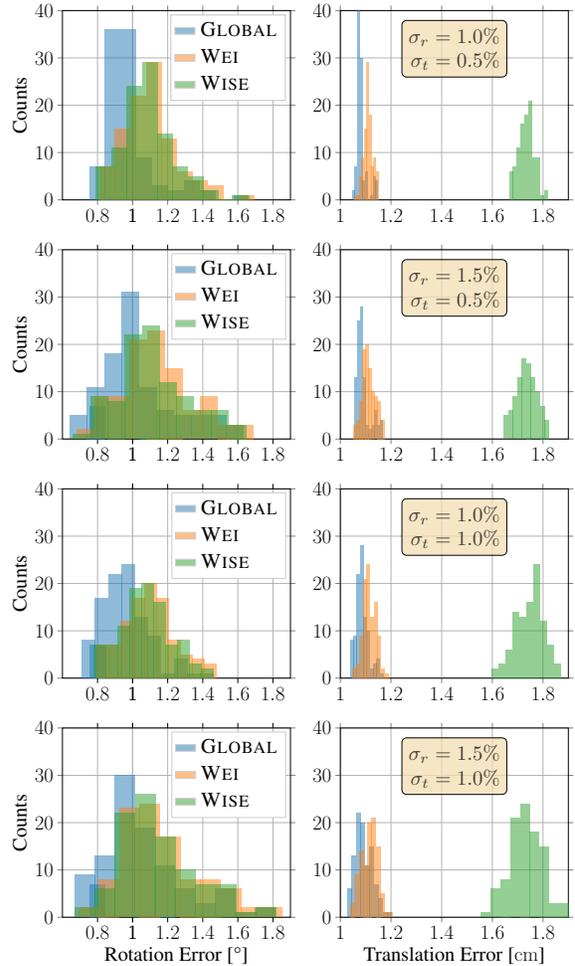}
    }
    \caption{Results for the Brookshire dataset~\cite{brookshire2013extrinsic} with artificial noise are shown.
    While both DQ approaches perform better with respect to the translation compared to the HM approach, our \method{Global} approach shows the smallest errors in rotation and translation.}
    \label{fig:noised_brookshire}
\end{figure}

As a next step, we evaluate the calibration performance with respect to noise on the Brookshire dataset~\cite{brookshire2013extrinsic} with artificial noise, similar to the evaluation in~\cite{wise2020certifiably}.
Our \method{Global} method is compared to \method{Wei} and \method{Wise}, as they are the fastest and most stable competitors based on the results in Section~\ref{sec:artScale}.
Using the same noise parameters as in~\cite{wise2020certifiably}, we created 100 datasets with artificial noise. 
The results of the algorithms are presented in Figure~\ref{fig:noised_brookshire}.

First, the evaluation shows that the translation error distribution for \method{Wise} is shifted to the right for all noise parameters.
Thus, both DQ-based approaches outperform the HM-based approach with respect to the translation. 
In general, the variations of all compared methods increase for higher noise levels.
However, compared to the other approaches, \method{Global} is the most noise-robust approach, having the smallest errors for rotation and translation.
This confirms the results from the previous subsection.

\subsection{Monocular Camera Data}
\label{sec:eval_euroc}

Finally, the EuRoC MAV dataset~\cite{burri2016euroc} is used for a comparison on real-world monocular camera data.
It contains data of a drone flying in a machine hall, equipped with two monocular cameras and an IMU sensor.
Its position is tracked using a laser tracker.
Further, ground-truth poses with positions from the laser tracker and orientations estimated from the IMU measurements are provided.
We used these poses to evaluate the calibration estimation between the first camera and the drone origin.
The monocular camera motion was estimated using the open source real-time SLAM library ORB-SLAM2 \cite{mur2015orb}.
We evaluated and compared all approaches based on the machine hall recordings MH1, MH2, and MH3.
MH1 and MH2 are rated with easy difficulty, whereas MH3 is of medium difficulty for visual-inertial odometry.
The results are shown in Table~\ref{tab:euroc}.

Since the scaling strongly depends on the initialization of the algorithm used for monocular motion estimation, no ground-truth information for scaling is available.
However, the results in Section~\ref{sec:artScale} indicate that the scaling is appropriately estimated when transformation errors are small.

For all recordings, our \method{Fast} approach always yields the same solutions as our \method{Global} approach.
Except for a few outliers of \method{Schmidt} and \method{Wei}, rotation and translation errors for all compared approaches are in an equal order of magnitude.
However, the biggest difference between all approaches is the execution time.
Here, the methods of Schmidt et al.~\cite{schmidt2005calibration} cannot keep up with the other approaches, taking between $\SI{1}{\second}$ and $\SI{1.5}{\second}$.
The execution time of \method{Wise} is always around $\SI{25}{\milli\second}$ to $\SI{27}{\milli\second}$.
Our \method{Fast} approach outperforms all other methods with an execution time of about $\SI{2}{\milli\second}$, which is approximately $10$ times faster than \method{Wise} and about $5$ times faster than our \method{Global} method and \method{Wei}.

\begin{table*}[t]
    \begin{center}
        \setlength{\tabcolsep}{0.37em}
        \begin{tabular}{lO{2.2}O{1.3}O{1.3}O{4.2}|O{2.2}O{1.3}O{1.3}O{4.2}|O{2.2}O{1.3}O{1.3}O{4.2}}
    \toprule
    \textbf{Method} & \multicolumn{4}{c}{\textbf{MH1}} &
                        \multicolumn{4}{c}{\textbf{MH2}} &
                        \multicolumn{4}{c}{\textbf{MH3}} \\
    & $\varepsilon_t$ [$\si{\centi\metre}$] & $\varepsilon_r$ [$\si{\degree}$] & $\hat{\alpha}$ &{$t$ [$\si{\milli\second}$]}
    & $\varepsilon_t$ [$\si{\centi\metre}$] & $\varepsilon_r$ [$\si{\degree}$] & $\hat{\alpha}$ &{$t$ [$\si{\milli\second}$]}
    & $\varepsilon_t$ [$\si{\centi\metre}$] & $\varepsilon_r$ [$\si{\degree}$] & $\hat{\alpha}$ &{$t$ [$\si{\milli\second}$]}\\
    
    \midrule
    
    \method{Schmidt HM}~\cite{schmidt2005calibration}
        & 1.55 & 0.328 & 4.854 & 1429.10
        & \B 0.50 & 0.168 & 0.935 & 1310.42
        & \B 1.08 & 0.926 & 2.184 & 1131.81\\
    
    \method{Schmidt DQ}~\cite{schmidt2005calibration}
        & 2.60 & \B 0.190 & 4.823 & 1291.23
        & 0.65 & 0.168 & 0.935 & 1310.95
        & 1.14 & 0.926 & 2.183 & 1178.48\\
    
    \method{Wei}~\cite{wei2018calibration}
        & \B 1.50 & 0.241 & 4.854 & 9.70
        & 0.65 & \B 0.114 & 0.935 & 9.90 
        & 4.82 & \B 0.150 & 2.183 & 9.10\\
        
    \method{Wise}~\cite{wise2020certifiably}
        & 1.62 & 0.268 & 4.854 & 25.49   
        & 0.68 & 0.157 & 0.935 & 26.37  
        & 1.21 & 0.895 & 2.184 & 27.16\\ 
        
    \method{Fast} (ours)
        & 1.53 & 0.280 & 4.854 & \B 2.34
        & 0.62 & 0.158 & 0.935 & \B 2.09
        & \B 1.08 & 0.849 & 2.184 & \B 2.37\\
        
    \method{Global} (ours)
        & 1.53 & 0.281 & 4.854 & 9.21 
        & 0.62 & 0.158 & 0.935 & 8.30 
        & \B 1.08 & 0.849 & 2.184 & 8.80\\

    \bottomrule    
\end{tabular}
    \end{center}
    \caption{
        Results on the EuRoC MAV dataset~\cite{burri2016euroc} are presented. 
        Since no ground-truth for $\alpha$ is available, the respective optimization results are given without a reference.
    }
    \label{tab:euroc}
\end{table*}

\begin{table}[t!]
    \begin{center}
        \setlength{\tabcolsep}{0.37em}
        \begin{tabular}{lO{2.2}O{1.3}|O{1.3}O{1.3}O{1.3}|O{2.3}}
    \toprule
    \textbf{Method} & 
    $\varepsilon_t$ [$\si{\centi\metre}$] & $\varepsilon_r$ [$\si{\degree}$] & $\hat{\alpha}_{1}$ & $\hat{\alpha}_{2}$ & $\hat{\alpha}_{3}$ & {$t$ [$\si{\milli\second}$]}\\
    \midrule

    \method{Fast} & 1.13 & 0.627 & 4.856 & 0.935 & 2.181 & 3.76 \\
    \method{Global} & 1.13 & 0.626 & 4.856 & 0.935 & 2.181 & 18.29 \\
    
    \bottomrule    
\end{tabular}
    \end{center}
    \caption{
        Results for the multi-scale calibration on the EuRoC MAV dataset~\cite{burri2016euroc} using our approaches are presented.}
    \label{tab:multi_scale}
\end{table}

\subsection{Multi-Scale Calibration}

For evaluating our multi-scale calibration, we used the same machine hall datasets as in Section~\ref{sec:eval_euroc} since all were recorded with the same sensor setup and calibration.
As shown with the results in the previous subsection, each camera motion is scaled differently.
Of all compared approaches, only our methods support multiple scalings.

The results of the multi-scale calibration with our approaches are presented in Table~\ref{tab:multi_scale}.
Comparing the resulting scalings with the respective results in Section~\ref{sec:eval_euroc} shows that each $\alpha$ is still estimated precisely.
Furthermore, errors in rotation and translation are in the same range as the errors for each individual calibration.
At the same time, both approaches take less time than the sum of all execution times in Section~\ref{sec:eval_euroc}.
This shows that our multi-scale calibration is able to simultaneously take multiple datasets into account while being efficient and preserving individual scaling factors at the same time.

\section{Conclusion}

We have described a QCQP formulation for monocular hand-eye calibration based on dual quaternions that allows for combining multiple datasets with different translation scalings.
With this formulation, a fast local and a globally optimal solving approach from our previous work~\cite{horn2021online} could be applied.
The evaluation on simulated and real-world data showed that our problem formulation yields consistently low execution times while retaining high estimation accuracy.  
Especially the fast solving approach has the lowest run-time of all compared approaches and still yields globally optimal results in almost all cases.
Thus, our proposed methods successfully fill the gap between a globally optimal and an efficient dual quaternion based monocular hand-eye calibration.
For future work, it would be interesting to deduce the cause of the higher noise sensitivity for the scaled sensor in all algorithms.

\newpage

{\small
\bibliographystyle{ieee_fullname}
\bibliography{mybibfile}
}



\section*{Supplementary material}

# Supplementary material for
## *Globally Optimal Multi-Scale Monocular Hand-Eye Calibration Using Dual Quaternions*

Thomas Wodtko*  Markus Horn*  Michael Buchholz  Klaus Dietmayer
Ulm University, Institute of Measurement, Control and Microtechnology
Albert-Einstein-Allee 41, 89081 Ulm, Germany
{firstname}.{lastname}@uni-ulm.de

## 1. Substitution Constraints

In this section, we describe the additional constraints necessary in order to retain the substitution $s = \alpha\,r$ in more detail. For every valid pair of $r$ and $s$, there must exist an $\alpha$ for which $s_i = \alpha\,r_i, \forall i \in \{1, \ldots, 4\}$. This leads to the following conditions:

$$\frac{s_i}{r_i} = \frac{s_j}{r_j}, \quad \text{if } r_i \neq 0 \wedge r_j \neq 0, \tag{1a}$$

$$s_i = 0, \quad \text{if } r_i = 0. \tag{1b}$$

Geometrically described, this means that $r$ and $s$ have to be parallel. In three-dimensional space, this can be achieved by enforcing the cross product of both vectors to be a null vector. However, the cross product is not defined in four-dimensional space. Nevertheless, we can achieve the same result in our case with the following six constraints:

$$r_i s_j - r_j s_i = 0, \quad \text{with } (i, j) \in \{1, \ldots, 4\}^2, \, i < j. \tag{2}$$

Table 1 shows the influence of each constraint. It shows that if and only if the constraints (2) are enforced during optimization, the conditions (1) hold for all feasible values of $r$. In our case, $r = 0$ is infeasible since it is prevented by the additional constraint $\|r\|_2 = 1$.

Further, Table 1 also shows that even without the last three constraints, all conditions (1) are still met if $r_1 \neq 0$. Following from the definition of rotation quaternions, $r_1 = 0$ only occurs for a rotation angle between both sensors of exactly $180°$. In practice, this is highly unlikely due to numerical inaccuracies and noise within the input data. Thus, we decided to omit these constraints in our implementation in order to reduce the computational complexity.

| Rotation | | | | Constraint | | | | | |
| --- | --- | --- | --- | --- | --- | --- | --- | --- | --- |
| $r_1$ | $r_2$ | $r_3$ | $r_4$ | $r_1 s_2 = r_2 s_1$ | $r_1 s_3 = r_3 s_1$ | $r_1 s_4 = r_4 s_1$ | $r_2 s_3 = r_3 s_2$ | $r_2 s_4 = r_4 s_2$ | $r_3 s_4 = r_4 s_3$ |
| 0 | 0 | 0 | 0 | – | – | – | – | – | – |
| 0 | 0 | 0 | $\pm$ | – | – | $s_1 = 0$ | – | $s_2 = 0$ | $s_3 = 0$ |
| 0 | 0 | $\pm$ | 0 | – | $s_1 = 0$ | – | $s_2 = 0$ | – | $s_4 = 0$ |
| 0 | 0 | $\pm$ | $\pm$ | – | $s_1 = 0$ | $s_1 = 0$ | $s_2 = 0$ | $s_2 = 0$ | $\frac{s_3}{r_3} = \frac{s_4}{r_4}$ |
| 0 | $\pm$ | 0 | 0 | $s_1 = 0$ | – | – | $s_3 = 0$ | $s_4 = 0$ | – |
| 0 | $\pm$ | 0 | $\pm$ | $s_1 = 0$ | – | $s_1 = 0$ | $s_3 = 0$ | $\frac{s_2}{r_2} = \frac{s_4}{r_4}$ | $s_3 = 0$ |
| 0 | $\pm$ | $\pm$ | 0 | $s_1 = 0$ | $s_1 = 0$ | – | $\frac{s_2}{r_2} = \frac{s_3}{r_3}$ | $s_4 = 0$ | $s_4 = 0$ |
| 0 | $\pm$ | $\pm$ | $\pm$ | $s_1 = 0$ | $s_1 = 0$ | $s_1 = 0$ | $\frac{s_2}{r_2} = \frac{s_3}{r_3}$ | $\frac{s_2}{r_2} = \frac{s_4}{r_4}$ | $\frac{s_3}{r_3} = \frac{s_4}{r_4}$ |
| $\pm$ | 0 | 0 | 0 | $s_2 = 0$ | $s_3 = 0$ | $s_4 = 0$ | – | – | – |
| $\pm$ | 0 | 0 | $\pm$ | $s_2 = 0$ | $s_3 = 0$ | $\frac{s_1}{r_1} = \frac{s_4}{r_4}$ | – | $s_2 = 0$ | $s_3 = 0$ |
| $\pm$ | 0 | $\pm$ | 0 | $s_2 = 0$ | $\frac{s_1}{r_1} = \frac{s_3}{r_3}$ | $s_4 = 0$ | $s_2 = 0$ | – | $s_4 = 0$ |
| $\pm$ | 0 | $\pm$ | $\pm$ | $s_2 = 0$ | $\frac{s_1}{r_1} = \frac{s_3}{r_3}$ | $\frac{s_1}{r_1} = \frac{s_4}{r_4}$ | $s_2 = 0$ | $s_2 = 0$ | $\frac{s_3}{r_3} = \frac{s_4}{r_4}$ |
| $\pm$ | $\pm$ | 0 | 0 | $\frac{s_1}{r_1} = \frac{s_2}{r_2}$ | $s_3 = 0$ | $s_4 = 0$ | $s_3 = 0$ | $s_4 = 0$ | – |
| $\pm$ | $\pm$ | 0 | $\pm$ | $\frac{s_1}{r_1} = \frac{s_2}{r_2}$ | $s_3 = 0$ | $\frac{s_1}{r_1} = \frac{s_4}{r_4}$ | $s_3 = 0$ | $\frac{s_2}{r_2} = \frac{s_4}{r_4}$ | $s_3 = 0$ |
| $\pm$ | $\pm$ | $\pm$ | 0 | $\frac{s_1}{r_1} = \frac{s_2}{r_2}$ | $\frac{s_1}{r_1} = \frac{s_3}{r_3}$ | $s_4 = 0$ | $\frac{s_2}{r_2} = \frac{s_3}{r_3}$ | $s_4 = 0$ | $s_4 = 0$ |
| $\pm$ | $\pm$ | $\pm$ | $\pm$ | $\frac{s_1}{r_1} = \frac{s_2}{r_2}$ | $\frac{s_1}{r_1} = \frac{s_3}{r_3}$ | $\frac{s_1}{r_1} = \frac{s_4}{r_4}$ | $\frac{s_2}{r_2} = \frac{s_3}{r_3}$ | $\frac{s_2}{r_2} = \frac{s_4}{r_4}$ | $\frac{s_3}{r_3} = \frac{s_4}{r_4}$ |

Table 1. The individual influence of each constraint in (2) is presented. For the rotation entries $r_i$, 0 denotes $r_i = 0$ and $\pm$ denotes $r_i \neq 0$. The table shows that the conditions (1) are met for all feasible values of $r$ since $r = 0$ is infeasible.



\end{document}